\documentclass{article}
\usepackage{ijcai25}

\usepackage{times}
\usepackage{soul}
\usepackage{url}
\usepackage[hidelinks]{hyperref}
\usepackage[utf8]{inputenc}
\usepackage[small]{caption}
\usepackage{graphicx}
\usepackage{amsmath}
\usepackage{amsthm}
\usepackage{booktabs}
\usepackage{algorithm}
\usepackage{algorithmic}
\usepackage[switch]{lineno}

\usepackage{tikz}
\usepackage{tikz-qtree}
\usepackage{Figures/tikzstyle}

\definecolor{lightblue}{RGB}{138,160,205}

\title{Reasoning RAG via System 1 or System 2: A Survey on Reasoning Agentic Retrieval-Augmented Generation for Industry Challenges}

\author{
Jintao Liang$^1$
\and
Gang Su$^2$\and
Huifeng Lin$^3$\and 
You Wu$^3$\and
Rui Zhao$^{5,6}$\And
Ziyue Li$^4$\thanks{Corresponding author.} \\
\affiliations
$^1$Beijing University of Posts and Telecommunications\\
$^2$University of Georgia\\
$^3$South China University of Technology\\
$^4$Technical University of Munich, University of Cologne\\
$^5$SenseTime Research\\
$^6$Qingyuan Research Institute, Shanghai Jiaotong University\\
\emails
ljt2021@bupt.edu.cn,
\{gangsuedu,huifeng.work,wuyouscut\}@gmail.com,
zhaorui@sensetime.com,
zlibn@wiso.uni-koeln.de
}

\begin{document}

\maketitle

\begin{abstract}
Retrieval-Augmented Generation (RAG) has emerged as a powerful framework to overcome the knowledge limitations of Large Language Models (LLMs) by integrating external retrieval with language generation. While early RAG systems based on static pipelines have shown effectiveness in well-structured tasks, they struggle in real-world scenarios requiring complex reasoning, dynamic retrieval, and multi-modal integration. To address these challenges, the field has shifted toward \textit{Reasoning Agentic RAG}, a paradigm that embeds decision-making and adaptive tool use directly into the retrieval process. In this paper, we present a comprehensive review of Reasoning Agentic RAG methods, categorizing them into two primary systems: \textit{predefined reasoning}, which follow fixed modular pipelines to boost reasoning, and \textit{agentic reasoning}, where the model autonomously orchestrates tool interaction during inference. We analyze representative techniques under both paradigms, covering architectural design, reasoning strategies, and tool coordination. Finally, we discuss key research challenges and propose future directions to advance the flexibility, robustness, and applicability of reasoning agentic RAG systems. Our collection of the relevant researches  has been  organized into a \href{https://github.com/ByebyeMonica/Reasoning-Agentic-RAG}{\textcolor{blue}{Github Repository}}.


\end{abstract}

\section{Introduction}
Large Language Models (LLMs)~\cite{singh2023exploring,zhao2023survey,zhu2024large} have demonstrated remarkable capabilities in natural language understanding and generation, enabling a wide array of applications from open-domain question-answer (QA) to task-specific dialogue systems. 
However, LLMs rely on static training data, making them prone to hallucinations and limiting their ability to provide accurate, up-to-date information in dynamic or knowledge-intensive tasks~\cite{rawte2023troubling,zhang2023siren,huang2025survey}.
Retrieval-Augmented Generation (RAG)~\cite{chen2024benchmarking,lewis2020retrieval,gao2023retrieval} has attracted significant attention as a promising approach to overcome the knowledge limitations of LLMs resulting from static pretraining. 
By integrating relevant information from external knowledge bases or search engines, RAG enhances factual accuracy and broadens the model's temporal and domain coverage \cite{zhao2024retrieval,li2024enhancing}. 
Traditional RAG methods have demonstrated strong performance when queries are well-formed and the necessary information is readily available in the retrieved context.

Despite the effectiveness of basic RAG methods, they often struggle when applied to real-world, industrial-scale applications involving complex and heterogeneous data. 
For example, in multi-document scenarios, relevant information is spread across sources, requiring not just retrieval but also coherent synthesis~\cite{wang2025retrieval,wang2024knowledge}. 
Naively concatenating retrieved passages can lead to fragmented or contradictory responses, particularly in domains like legal or biomedical QA where multi-hop reasoning is critical.
Additionally, most RAG systems are limited to text-only processing and cannot natively handle multi-modal inputs such as tables, charts, or images~\cite{ma2024multi,yu2025mramg}.
This limits their ability to operate in data-rich environments like enterprise intelligence, scientific reporting, or technical support, where visual and structured data play a central role \cite{lin2023fine,yu2024visrag}.

\begin{figure}[t]
  \centering
  \includegraphics[width=1\linewidth]{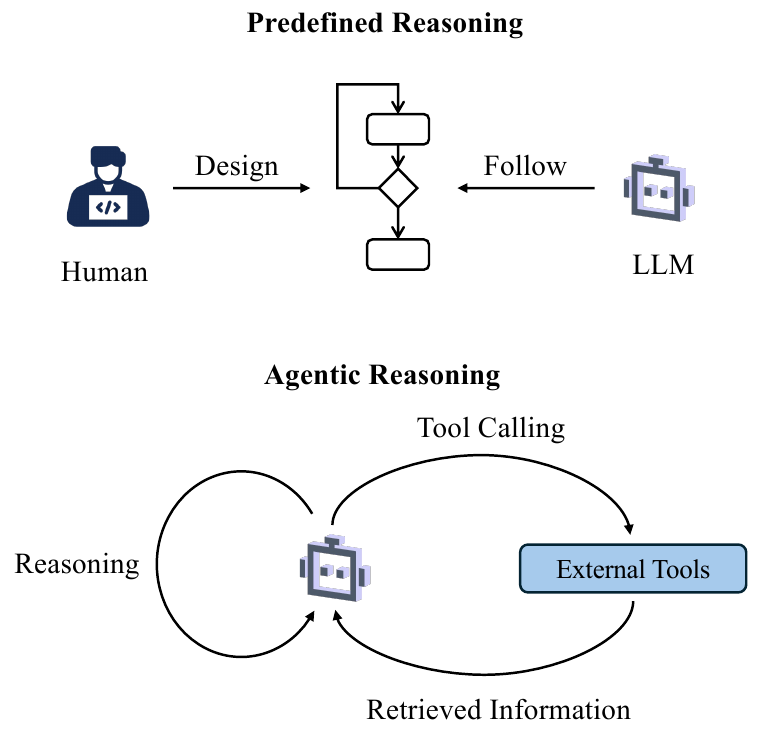}
  \caption{Overview of two major types of reasoning Agentic Systems.}
  \label{high_level_description}
\end{figure}

To address these limitations of basic RAG in handling complex, real-world tasks, recent research has turned to \textit{Agentic RAG} \cite{ravuru2024agentic}, a paradigm that tightly integrates retrieval with reasoning and decision-making. 
Unlike static pipelines, Agentic RAG treats retrieval not as a one-off preprocessing step, but as a dynamic, context-sensitive operation guided by the model's ongoing reasoning process. This reasoning-centric perspective is crucial for applications that demand multi-step problem solving, adaptive information acquisition, and tool-assisted synthesis. 
Within this paradigm, as shown in Figure ~\ref{high_level_description}, two major types of reasoning agentic systems have emerged based on how control and decision-making are handled: \textit{predefined reasoning}, which follow structured, rule-based plans with fixed pipelines to boost reasoning for retrieval and generation; and \textit{agentic reasoning}, where the model actively monitors its reasoning process and determines when and how to retrieve or interact with external tools.
These two workflows form the basis of \textit{Reasoning Agentic RAG}, which unifies structured and autonomous approaches for more intelligent, context-aware retrieval-augmented reasoning.

    \textbf{Predefined reasoning} adopts structured and modular RAG pipelines where the retrieval and reasoning steps are explicitly designed, following fixed control pipeline. 
    These workflows typically decompose tasks into discrete components such as query reformulation, document retrieval, re-ranking, and answer synthesis, executed in a linear or orchestrated fashion. 
    In general, predefined reasoning spans several architectural variants: 
\textit{route-based} methods selectively trigger retrieval based on context or model uncertainty, such as low confidence scores or ambiguous intermediate outputs~\cite{wang2024adaptiveretrievalaugmentedgenerationconversational}; \textit{loop-based} methods enable limited iteration through retrieval-feedback cycles, supporting multiple rounds of refinement \cite{asai2023selfraglearningretrievegenerate,yang2024crag};
\textit{tree-based} methods organize information hierarchically to support structured exploration \cite{sarthi2024raptorrecursiveabstractiveprocessing,hu2025mctsragenhancingretrievalaugmentedgeneration}; and \textit{hybrid-modular} frameworks compose specialized modules into a flexible but still rule-driven workflow \cite{jeong2024adaptiveraglearningadaptretrievalaugmented,gao2024modularragtransformingrag}. 
    These workflows prioritize control and modularity, making them suitable for tasks requiring efficient computation and customization. However, their reasoning remains constrained by predesigned execution paths, limiting flexibility in evolving and open-ended tasks.

    \textbf{Agentic reasoning} repositions the LLM as an active decision maker, that autonomously orchestrates retrieval and tool use throughout the reasoning process. 
    Instead of executing a fixed plan, the model identifies knowledge gaps, formulates queries, retrieves external information via tools such as search engines or APIs, and integrates the retrieved contents into an evolving solution.
    This dynamic interplay of reasoning and tool use enables the system to tackle complex, multi-turn tasks that require iterative refinement and adaptive information synthesis.
    There are two primary methods for implementing agentic reasoning. 
    The first is \textit{prompt-based} methods, which leverages the in-context reasoning and instruction-following capabilities of pretrained LLMs~\cite{yao2023reactsynergizingreasoningacting,press2023measuringnarrowingcompositionalitygap,li2025searcho1agenticsearchenhancedlarge}. In this setting, the model is guided by carefully crafted prompts or embedded control tokens that instruct it when to retrieve, what actions to take, and how to integrate external information. These methods require no additional training, making them lightweight and adaptable across tasks. 
    The second paradigm is \textit{training-based} methods, where models are explicitly optimized through reinforcement learning, to determine when and how to invoke external tools~\cite{jiang2025deepretrievalhackingrealsearch,jin2025searchr1trainingllmsreason,zheng2025deepresearcherscalingdeepresearch}. 
    This paradigm enables more fine-grained and strategic tool usage, enabling models to learn long-term planning and develop retrieval policies tailored to complex tasks. 
    Owing to its autonomy and adaptability, agentic reasoning has shown strong performance in open-domain QA, scientific reasoning, and multi-stage decision-making scenarios.

\begin{table*}[t]
\centering
\begin{tabular}{@{}lll@{}}
\toprule
\textbf{System Type} & \textbf{Reasoning Workflow}       & \textbf{Description}                                 \\ \midrule
System 1             & Predefined Reasoning& Structured, modular, rule-based execution. \\
System 2            & Agentic Reasoning & Autonomous, adaptive, model-driven decision-making. \\ \bottomrule
\end{tabular}
\caption{Cognitive system alignment of reasoning workflows.}
\label{tab:cognitive_alignment}
\end{table*}

\textbf{Perspective of Cognitive Science - System 1 and System 2:} To further contextualize predefined and agentic reasoning within the dual-process theory of cognition—commonly referred to as System 1 and System 2 thinking \cite{yang2024llm2,li2025system} — we can draw an analogy between these RAG paradigms and human cognitive modes. \begin{itemize}
    \item Predefined reasoning resembles System 1 thinking: fast, structured, and efficient, relying on predefined heuristics and modular workflows that mirror habitual or rule-based cognition. While this enables rapid execution and predictable behavior, it often lacks the flexibility to adapt beyond its design.
    \item In contrast, agentic reasoning aligns more closely with System 2 thinking: slow, deliberative, and adaptive. Here, the LLM actively engages in reasoning, planning, and decision-making, dynamically leveraging external tools and retrieved knowledge to address complex, novel tasks. This reflective mode allows the model to identify gaps, reassess strategies, and adjust its behavior—traits characteristic of conscious, analytical human reasoning.
\end{itemize}  

By framing these paradigms through the lens of cognitive systems, we highlight the trade-off between efficiency and adaptability, and the growing capacity of agentic RAG to emulate more sophisticated, human-like problem solving. Table~\ref{tab:cognitive_alignment} aligns predefined and agentic reasoning with the dual-system theory from cognitive science, illustrating their respective control structures and behavioral characteristics.

The paper systematically reviews and analyzes the current research approaches and future development paths of Reasoning Agentic RAG, summarizing them into two primary technical paradigms. 
The remainder of the paper is organized as follows: Section~\ref{related work} introduces related work; Section~\ref{predefined} and Section~\ref{agentic} dive into the two types of reasoning workflows within Agentic RAG, predefined reasoning and agentic reasoning, respectively. 
Section~\ref{future} outlines future research directions, and Section~\ref{conculsion} concludes the paper.

\begin{figure}[h]
	\centering
	\includegraphics[width=1\linewidth]{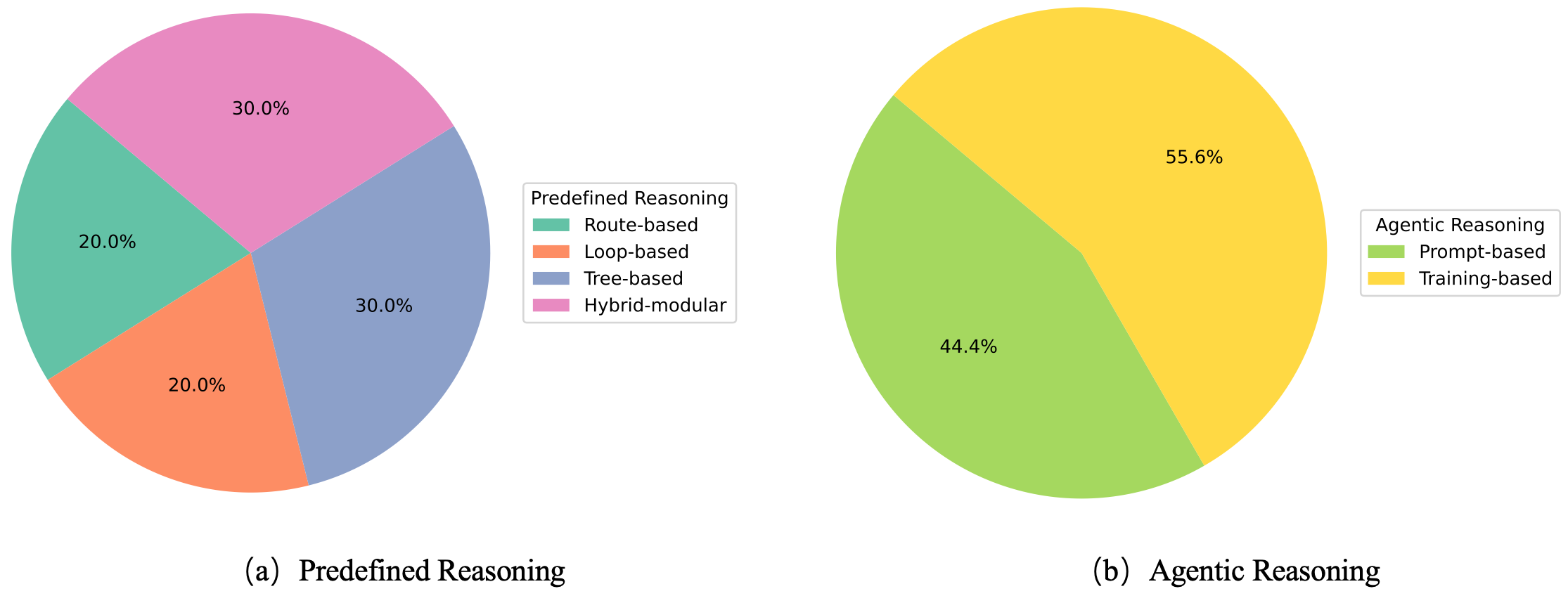}
	\caption{Distributed Works of Reasoning Agentic RAG.}
	\label{reasoning type} 
\end{figure}

\section{Related Work}
\label{related work}
\subsection{Basic RAG}

\textit{Retrieval-Augmented Generation} (RAG) was introduced to overcome the static knowledge limitations of LLMs by integrating external retrieval mechanisms during inference~\cite{chen2024benchmarking,gao2023retrieval}. Naive RAG methods represent the earliest implementations, typically using sparse retrieval techniques like BM25~\cite{robertson2009probabilistic} to fetch documents based on keyword overlap~\cite{ma2023query}. While efficient for simple factoid queries, these approaches offered limited semantic understanding, thus often retrieving noisy or redundant content and failing to reason across multiple sources.


The emergence of Advanced RAG and Modular RAG was aimed at addressing key limitations of the Naive RAG, particularly in terms of retrieval precision, information integration, and system flexibility~\cite{gao2023retrieval}. 
Advanced RAG improves retrieval quality through techniques such as dense semantic matching, re-ranking, and multi-hop querying, while also introducing refined indexing strategies like fine-grained chunking and metadata-aware retrieval. 
Modular RAG rethinks the Naive RAG by breaking down the end-to-end process of indexing, retrieval, and generation into discrete, configurable modules.
This design allows for greater architectural flexibility and enables system developers to incorporate diverse techniques into specific stages, such as enhancing retrieval with fine-tuned search modules \cite{lin2023ra}.
In response to specific task demands, various restructured and iterative module designs have also emerged. As a result, modular RAG has increasingly become a dominant paradigm in the field, supporting both serialized pipeline execution and end-to-end learning across modular components.

Despite their effectiveness, basic RAG workflows are limited by static control logic and lack the ability to reflect, adapt, or assess the sufficiency of retrieved information. These constraints reduce their suitability for tasks requiring iterative reasoning, tool use, or multi-modal integration. Thus, Agentic RAG has proposed to embed reasoning and decision-making into the retrieval process. This work focuses on reasoning Agentic RAG approaches that enable more autonomous and context-aware information processing.

\tikzstyle{leaf}=[draw=hiddendraw,
    rounded corners,
    minimum height=1em,
    text opacity=1, 
    align=center,
    fill opacity=.5,  
    text=black,
    align=left,
    font=\scriptsize,
    inner xsep=3pt,
    inner ysep=1pt,
    ]
\tikzstyle{middle}=[draw=hiddendraw,
    rounded corners,
    minimum height=1em,
    color = lightblue,
    fill=lightblue!40, 
    text opacity=1, 
    align=center,
    fill opacity=.5,  
    text=black,
    align=center,
    font=\scriptsize,
    inner xsep=3pt,
    inner ysep=1pt,
    ]
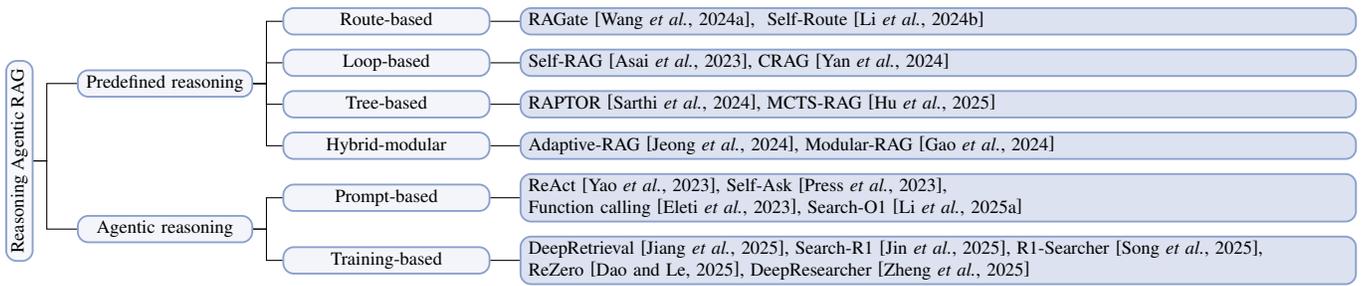
\begin{figure*}[t]
\centering
\begin{forest}
  for tree={
  forked edges,
  grow=east,
  reversed=true,
  anchor=base west,
  parent anchor=east,
  child anchor=west,
  base=middle,
  font=\scriptsize,
  rectangle,
  line width=0.7pt,
  edge={draw=black},
  draw=blue,
  rounded corners,align=left,
  minimum width=2em,
    s sep=5pt,
    inner xsep=3pt,
    inner ysep=1pt},
  where level=1{text width=4.5em}{},
  where level=2{text width=6em,font=\scriptsize}{},
  where level=3{font=\scriptsize}{},
  where level=4{font=\scriptsize}{},
  where level=5{font=\scriptsize}{},
  [Reasoning Agentic RAG, middle, rotate=90, anchor=north, draw=lightblue, fill=lightblue!15
        [Predefined reasoning, middle, text width=6em, color=lightblue, fill=lightblue!20, text=black
            [Route-based, middle, text width=7.2em, color=lightblue, fill=lightblue!20, text=black
                [RAGate \cite{wang2024adaptiveretrievalaugmentedgenerationconversational}{, }
                 Self-Route \cite{li2024retrievalaugmentedgenerationlongcontext}, 
                 leaf, text width=31em, color=lightblue, fill=lightblue!60, text=black]
            ]
            [Loop-based, middle, text width=7.2em, color=lightblue, fill=lightblue!20, text=black
                [Self-RAG \cite{asai2023selfraglearningretrievegenerate}{,} 
                 CRAG \cite{yan2024correctiveretrievalaugmentedgeneration},
                 leaf, text width=31em, color=lightblue, fill=lightblue!60, text=black]
            ]
            [Tree-based, middle, text width=7.2em, color=lightblue, fill=lightblue!20, text=black
                [RAPTOR \cite{sarthi2024raptorrecursiveabstractiveprocessing}{,}
                 MCTS-RAG \cite{hu2025mctsragenhancingretrievalaugmentedgeneration},
                 leaf, text width=31em, color=lightblue, fill=lightblue!60, text=black]
            ]
            [Hybrid-modular, middle, text width=7.2em, color=lightblue, fill=lightblue!20, text=black
                [Adaptive-RAG \cite{jeong2024adaptiveraglearningadaptretrievalaugmented}{,} 
                 Modular-RAG \cite{gao2024modularragtransformingrag}, 
                 leaf, text width=31em, color=lightblue, fill=lightblue!60, text=black]
            ]
        ]
        [Agentic reasoning, middle, text width=6em, color=lightblue, fill=lightblue!20, text=black
            [Prompt-based, 
            middle, text width=7.2em, color=lightblue, fill=lightblue!20, text=black
                [ReAct \cite{yao2023reactsynergizingreasoningacting}{,} Self-Ask \cite{press2023measuringnarrowingcompositionalitygap}{,}
                \\
                Function calling \cite{openai_function_calling}{,} Search-O1 \cite{li2025searcho1agenticsearchenhancedlarge}, 
                leaf, text width=31em, color=lightblue, fill=lightblue!60, text=black]
            ]
            [Training-based, middle, text width=7.2em, color=lightblue, fill=lightblue!20, text=black
                [DeepRetrieval \cite{jiang2025deepretrievalhackingrealsearch}{,} Search-R1 \cite{jin2025searchr1trainingllmsreason}{,} R1-Searcher \cite{song2025r1searcherincentivizingsearchcapability}{,}
                \\
                ReZero \cite{dao2025rezeroenhancingllmsearch}{,} DeepResearcher \cite{zheng2025deepresearcherscalingdeepresearch}, 
                leaf, text width=31em, color=lightblue, fill=lightblue!60, text=black]
            ]
        ]            
    ]
\end{forest}
\vspace{-5mm}
\caption{A taxonomy of Reasoning Agentic RAG.}

\label{fig:taxonomy}
\vspace{-4mm}
\end{figure*}

\subsection{Reasoning Agentic RAG}

\begin{figure*}[h]
  \centering
  \includegraphics[width=0.99\linewidth]{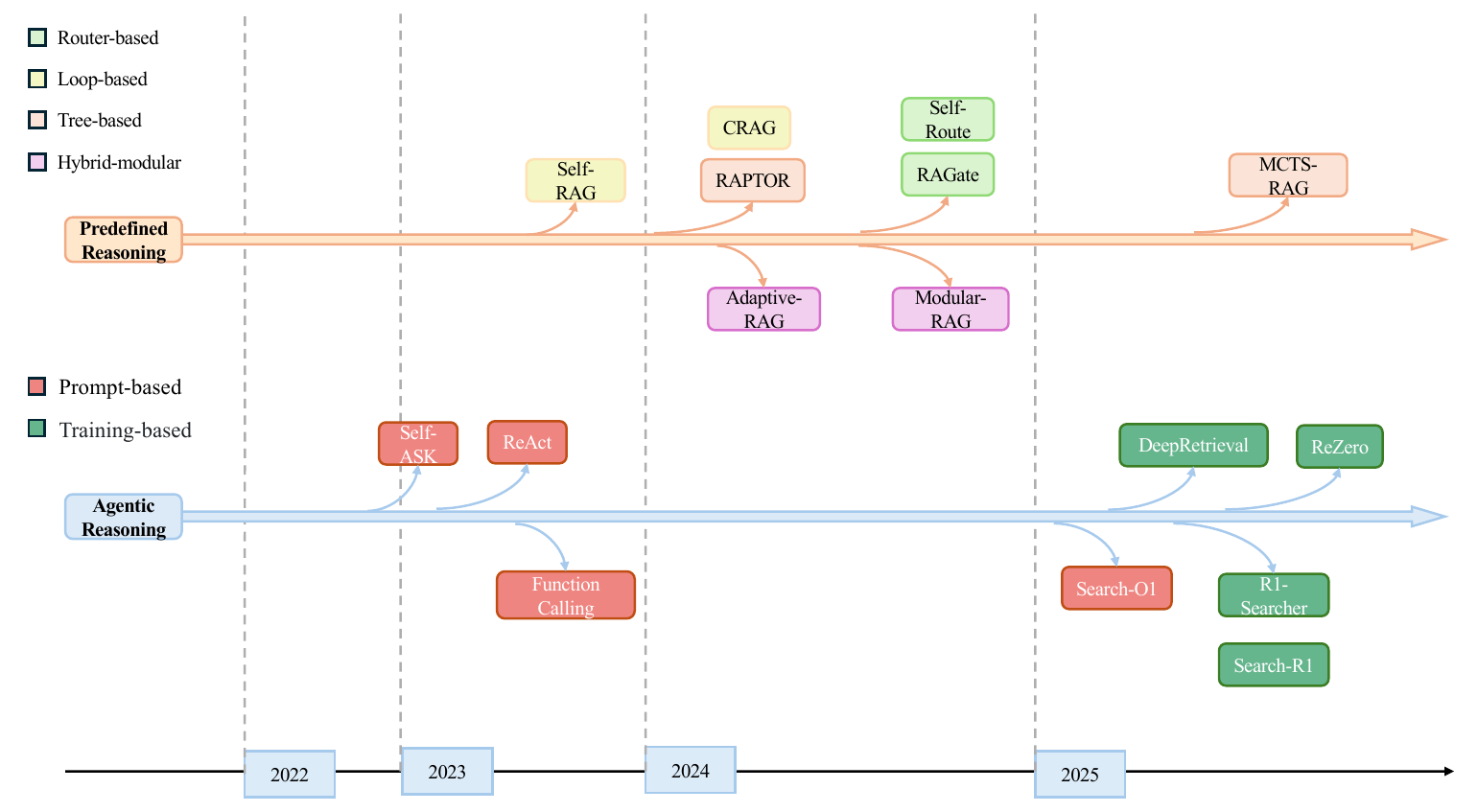}
  \caption{Illustration of the evolution of Reasoning Agentic RAG.}
  \label{timeline}
\end{figure*}

The year 2025 is marked as the year of agentic AI, with applications emerging such as agentic LLMs and so on \cite{ruan2023tptu,kong2024tptu,zhang2024controlling}. Recent advances in RAG have seen a shift from static, rule-driven retrieval pipelines toward dynamic, reasoning-driven architectures, collectively referred to as \textit{Reasoning Agentic RAG}. These systems embed decision-making into the retrieval process, enabling models to actively determine when, what, and how to retrieve based on their internal reasoning trajectory. 
As shown in Figure~\ref{fig:taxonomy}, Reasoning Agentic RAG approaches can be broadly categorized into two paradigms: \textit{predefined reasoning} and \textit{agentic reasoning}.

Predefined reasoning depends on structured, rule-based pipelines where the retrieval and reasoning stages are modularized and fixed in advance. These workflows often include components for query reformulation, document retrieval, re-ranking, and response generation, coordinated by static control logic. 
RAGate~\cite{wang2024adaptiveretrievalaugmentedgenerationconversational} exemplifies route-based designs, where retrieval is conditionally triggered based on the context or model confidence, enabling the system to skip unnecessary operations and focus on knowledge-intensive inputs. 
Self-RAG~\cite{asai2023selfraglearningretrievegenerate} introduces loop-based reasoning by enabling the model to self-reflect and iteratively refine its responses, while RAPTOR~\cite{sarthi2024raptorrecursiveabstractiveprocessing} leverages a recursive tree structure to hierarchically summarize and organize retrieved content, supporting multi-hop and abstractive reasoning. Building on these foundations, more advanced frameworks like Adaptive-RAG~\cite{jeong2024adaptiveraglearningadaptretrievalaugmented} combine dynamic routing and retrieval adaptation, enabling models to select optimal reasoning paths. Modular-RAG~\cite{gao2024modularragtransformingrag} extends this idea by dividing the RAG pipeline into interoperable modules like retrievers, rerankers and generators, which can be flexibly composed into hybrid workflows.
These designs enabling more flexible orchestration while still operating under predefined execution paths.

Agentic reasoning empowers the LLM to act as an autonomous agent, dynamically deciding how to interact with external tools based on its current reasoning state. These workflows tightly couple reasoning with tool use, enabling the model to issue retrieval queries, assess results, and iteratively adapt its actions. Two main implementation strategies have emerged: \textit{prompt-based} and \textit{training-based} approaches. 
Prompt-based methods leverage the instruction-following abilities of pretrained LLMs to drive agentic behavior without additional training. 
For example, ReAct~\cite{yao2023reactsynergizingreasoningacting} interleaves reasoning steps with tool use to guide retrieval based on emerging knowledge gaps. Other methods like Self-Ask~\cite{press2023measuringnarrowingcompositionalitygap} and Search-o1~\cite{li2025searcho1agenticsearchenhancedlarge} support decomposition into sub-questions or trigger retrieval mid-generation. 
Additionally, function calling mechanisms \cite{openai_function_calling} built into commercial LLMs such as GPT and Gemini offer structured interfaces for tool use, further enabling prompt-based agentic control.
In parallel, training-based approaches aim to explicitly teach LLMs to reason and retrieve in a unified, goal-driven manner by leveraging reinforcement learning (RL) to optimize tool-use behavior.
DeepRetrieval~\cite{jiang2025deepretrievalhackingrealsearch} trains models to reformulate queries by maximizing retrieval metrics. 
Search-R1~\cite{jin2025searchr1trainingllmsreason} and R1-Searcher~\cite{song2025r1searcherincentivizingsearchcapability} both adopt a two-stage, outcome-driven RL framework that enables LLMs to learn when and what to search within a reasoning trajectory.
ReZero~\cite{dao2025rezeroenhancingllmsearch}incentivizes persistence, rewarding effective retry strategies. DeepResearcher~\cite{zheng2025deepresearcherscalingdeepresearch} pushes further by training agents in open web environments, enabling robust search and synthesis across diverse, unstructured sources.

\section{Predefined Reasoning}
\label{predefined}



Agents and RAG are increasingly integrated in advanced AI systems. 
By augmenting LLMs with external knowledge retrieval, RAG enables agents to ground their reasoning in relevant information. 
In turn, agent-based reasoning which includes planning, tool use and self-reflection, enhances RAG by guiding the model on what information to retrieve and how to incorporate it into the reasoning process.
This synergy supports a predefined reasoning, where the agent iteratively queries external sources (e.g., a local database or web search) and refines its reasoning based on the retrieved evidence. We categorize predefined RAG reasoning workflows into four broad types based on their structural and reasoning characteristics as follows.


\textbf{Route-based Approaches:} RAG incorporates dynamic routing mechanisms that direct queries along different retrieval or reasoning paths based on predefined conditions—such as query type, model uncertainty, or confidence estimation—while still operating within a fixed architecture.  
RAGate \cite{wang2024adaptiveretrievalaugmentedgenerationconversational} uses the conversation context and model confidence to route only those dialogue turns that truly require external knowledge to a RAG process. This ensures the system can bypass retrieval for straightforward prompts while invoking it for knowledge-intensive queries, exemplifying conditional RAG in dialogue. Self-Route \cite{li2024retrievalaugmentedgenerationlongcontext} introduced dynamically routes queries to either RAG or Long-Context (LC) models based on the model’s confidence-based routing. This method significantly reduces computation cost while maintaining performance comparable to LC models. 

\textbf{Loop-based Approaches:} RAG operates within a feedback loop that supports multiple rounds of refinement. The system can self-reflect, critique intermediate outputs, and iteratively update retrieval inputs to improve generation quality.
Self-RAG \cite{asai2023selfraglearningretrievegenerate} is a foundational example of this controlled reasoning loop. In the Self-RAG workflow, a single LLM agent engages in self-reflection during generation to improve its output. Instead of relying on a fixed retrieved context, the model can decide mid-generation to fetch additional information or to critique its own draft answer. CRAG \cite{yan2024correctiveretrievalaugmentedgeneration} introduced loop-based corrective feedback mechanism into the retrieval process. In the CRAG workflow, a lightweight retrieval evaluator assigning the confidence scores about the quality of the retrieved chunks/documents — categorized as correct, incorrect, or ambiguous. When retrieval quality is deemed suboptimal, the system activates corrective strategies such as query rewriting or external web search to gather better evidence. The system refines the retrieved content into a focused context and iteratively improves retrieval until a satisfactory output is generated. 
\begin{figure}[!h]
	\centering
	\includegraphics[width=0.9\linewidth]{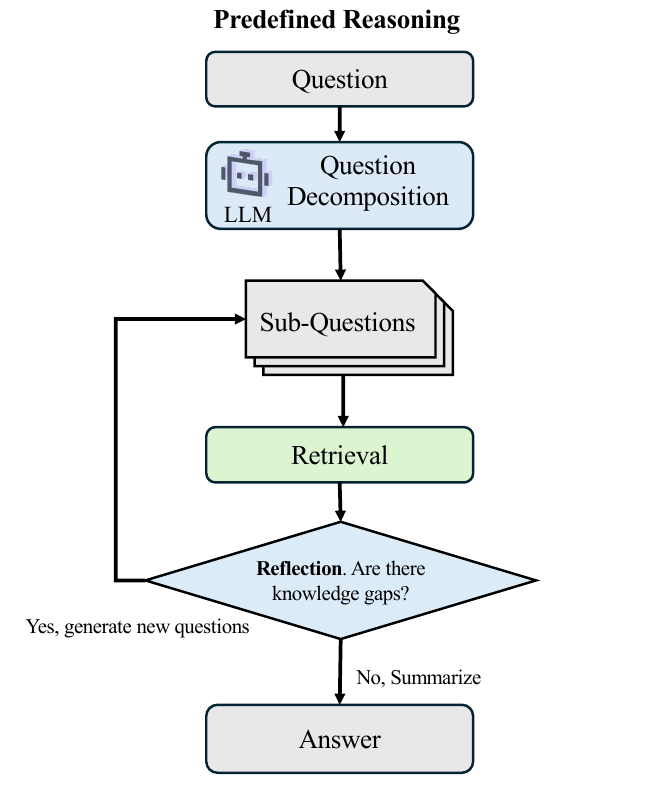}
	\caption{A demonstration of Predefined Reasoning.}
	\label{predefined reasoning figure} 
\end{figure}

\textbf{Tree-based Approaches:} RAG organizes the retrieval process hierarchically, often using recursive structures such as trees to support multi-hop reasoning or document summarization.
RAPTOR \cite{sarthi2024raptorrecursiveabstractiveprocessing} introduces a recursive tree structure from documents, allowing for more efficient and context-aware information retrieval. This approach enhances RAG by creating a summary tree from text chunks, providing deeper insights and overcoming limitations of short, contiguous text retrieval. MCTS-RAG \cite{hu2025mctsragenhancingretrievalaugmentedgeneration} integrates a Monte Carlo Tree Search loop into the RAG process for complex reasoning tasks. MCTS-RAG dynamically integrates retrieval and reasoning through an iterative decision-making process. Unlike standard RAG methods, which typically retrieve information independently from reasoning and thus integrate knowledge suboptimally, or conventional MCTS reasoning, which depends solely on internal model knowledge without external facts, MCTS-RAG combines structured reasoning with adaptive retrieval. 
\textbf{Hybrid-modular Approaches:} RAG in its most flexible form combines routing, looping, reflection, and modular orchestration. Tasks are divided among specialized components, coordinated by an agent that can dynamically reconfigure the workflow according to the query or reasoning context.  
Adaptive-RAG \cite{jeong2024adaptiveraglearningadaptretrievalaugmented} extends the Self-RAG framework by introducing routing mechanisms that enable dynamic path selection. In addition to allowing the model to interleave retrieval and generation steps, it equips the agent with a decision-making router that selects appropriate retrieval strategies or reasoning pathways based on the query characteristics or the agent’s own uncertainty. Rather than simply determining whether to retrieve more information, the agent can choose which retrieval method to apply, what type of information to prioritize, or which downstream modules to engage. Modular-RAG \cite{gao2024modularragtransformingrag} is the most advanced incarnation that transform RAG into a LEGO-like modular framework, breaking the RAG process into an orchestrated pipeline of specialized modules. Rather than a single agent handling everything, a Modular-RAG architecture compartmentalizes tasks, e.g., one module for query reformulation, one for document retrieval, another for ranking or filtering results, and another for answer synthesis – all chained together in a composable workflow. The pipeline is composed by an agent that coordinates modular components, each of which can be optimized or swapped independently.

This progression of predefine reasoning workflows reflects a broader shift from static retrieval pipelines to dynamic, agent-driven reasoning systems. Modern predefined reasoning increasingly integrates planning, tool use, and decision-making components that allow flexible orchestration of retrieval and reasoning strategies. Rather than predefining rigid retrieval steps, these systems empower agents to determine what information to seek, how to use it, and when to adapt their approach—marking a move toward more autonomous and intelligent knowledge integration. A summary of the representative research works and open-source industrial/enterprise implementations across these predefined RAG workflow types is provided in Table~\ref{tab:taxonomy}.

\begin{table*}[!t]
\resizebox{\linewidth}{!}{
\begin{tabular}{@{}llllllll@{}}

   \\
\multicolumn{5}{c}{\textbf{Predefined Reasoning}} \\ 
\midrule\midrule

\textbf{Approach} &
  \textbf{Strategy} &
  \textbf{Control Type} &
  \textbf{Reasoning Complexity} &
  \textbf{Code} \\ \midrule
RAGate \cite{wang2024adaptiveretrievalaugmentedgenerationconversational} &
  Route-based &
  Adaptive &
  Medium &
  \href{https://github.com/wangxieric/RAGate}{Link} \\
self-RAG \cite{asai2023selfraglearningretrievegenerate} &
  Loop-based &
  Agentic &
  Medium &
  \href{https://github.com/AkariAsai/self-rag}{Link} \\
CRAG \cite{yan2024correctiveretrievalaugmentedgeneration} &
  Loop-based &
  Adaptive &
  Medium &
  \href{https://github.com/HuskyInSalt/CRAG}{Link} \\
MCTS-RAG \cite{hu2025mctsragenhancingretrievalaugmentedgeneration} &
  Tree-based &
  Agentic &
  High &
  \href{https://github.com/yale-nlp/MCTS-RAG}{Link} \\
RAPTOR \cite{sarthi2024raptorrecursiveabstractiveprocessing} &
  Tree-based &
  Fixed &
  Medium &
  \href{https://github.com/parthsarthi03/raptor}{Link} \\
Adaptive-RAG \cite{jeong2024adaptiveraglearningadaptretrievalaugmented} &
  Hybrid-modular &
  Adaptive &
  Medium &
  \href{https://github.com/starsuzi/Adaptive-RAG}{Link} \\
Modular-RAG \cite{gao2024modularragtransformingrag} &
  Hybrid-modular &
  Fixed &
  Low &
  N/A \\

DeepSearcher &
  Industry &
  Adaptive &
  Medium &
  \href{https://github.com/zilliztech/deep-searcher}{Link} \\
RAGFlow &
  Industry &
  Adaptive &
  Medium &
  \href{https://github.com/infiniflow/ragflow}{Link} \\
Haystack &
  Industry &
  Adaptive &
  Medium &
  \href{https://github.com/deepset-ai/haystack}{Link} \\
Langchain-Chatchat &
  Industry &
  Adaptive/Agentic &
  Medium &
  \href{https://github.com/chatchat-space/Langchain-Chatchat}{Link} \\
LightRAG &
  Industry &
  Adaptive &
  Medium &
  \href{https://github.com/HKUDS/LightRAG}{Link} \\
R2R &
  Industry &
  Agentic &
  High &
  \href{https://github.com/SciPhi-AI/R2R}{Link} \\
FlashRAG &
  Industry &
  Adaptive &
  Medium &
  \href{https://github.com/RUC-NLPIR/FlashRAG}{Link} \\

\bottomrule
 &
   &
   &
   &
   &
   &
   &
   \\
\multicolumn{5}{c}{\textbf{Agentic Reasoning}} \\ 
\midrule\midrule
  \textbf{Approach} &
  \textbf{Strategy} &
  \textbf{Training environment} &
  \textbf{Reward design} &
  \textbf{Code} \\ 
  \midrule
    ReAct \cite{yao2023reactsynergizingreasoningacting} &
    Prompt-based &
    N/A &
    N/A &
    \href{https://github.com/ysymyth/ReAct}{Link} \\

    Self-Ask \cite{press2023measuringnarrowingcompositionalitygap} &
    Prompt-based &
    N/A &
    N/A &
    \href{https://github.com/ofirpress/self-ask}{Link} \\
    
    Funciton calling \cite{openai_function_calling} &
    Prompt-based &
    N/A &
    N/A &
    N/A \\

    Search-O1 \cite{li2025searcho1agenticsearchenhancedlarge} &
    Prompt-based &
    N/A &
    N/A &
    \href{https://github.com/sunnynexus/Search-o1}{Link} \\

    Search-R1 \cite{jin2025searchr1trainingllmsreason} &
    Training-based &
    Local retrieval system &
    Answer reward &
    \href{https://github.com/PeterGriffinJin/Search-R1}{Link} \\

    R1-Searcher \cite{song2025r1searcherincentivizingsearchcapability} &
    Training-based &
    Local retrieval system &
    Retrieval reward, format reward, answer reward &
    \href{https://github.com/RUCAIBox/R1-Searcher}{Link} \\

    ReZero \cite{dao2025rezeroenhancingllmsearch} &
    Training-based &
    Local retrieval system &
    Retrieval reward, format reward, answer reward, retry reward &
    \href{https://github.com/menloresearch/ReZero}{Link} \\
    
    DeepRetrieval \cite{jiang2025deepretrievalhackingrealsearch} &
    Training-based &
    Restricted real-world search engine &
    Retrieval reward, format reward &
    \href{https://github.com/pat-jj/DeepRetrieval}{Link} \\

    DeepResearcher \cite{zheng2025deepresearcherscalingdeepresearch} &
    Training-based &
    Real-world search engine &
    Format reward, answer reward &
    \href{https://github.com/GAIR-NLP/DeepResearcher}{Link}
    
\\ \bottomrule
\end{tabular}}
\vspace{-2mm} 
\caption{A summary of Reasoning agentic rag.}

\label{tab:taxonomy}
\vspace{-4mm} 
\end{table*}

\section{Agentic Reasoning}
\label{agentic}


Beyond the predefined reasoning mentioned above, a more dynamic paradigm has emerged: the \textit{Agentic Reasoning}. In this setting, the LLM serves as an autonomous agent that not only generates text, but also actively manages retrieval. With advances in reasoning and instruction-following capabilities, the model can identify knowledge gaps, determine when and what to retrieve, and interact with external tools such as search engines or APIs. 
This tight integration of reasoning and tool use enables iterative decision-making, enabling the system to refine its responses based on newly retrieved information. As a result, agentic reasoning supports more flexible and adaptive problem-solving, extending RAG beyond basic QA to complex tasks such as scientific inquiry, multi-step reasoning, and strategic decision-making.
Agentic reasoning approaches can be broadly categorized by how the LLM learns to use tools:
\begin{itemize}
    \item \textbf{Prompt-Based Approaches:} These methods leverage the instruction-following, in-context learning and reasoning capabilities of pretrained LLMs, guiding tool use through carefully crafted prompts or built-in functionalities without additional training.
    
    \item \textbf{Training-Based Approaches:} These methods involve explicitly training LLMs, typically via reinforcement learning, to learn when and how to interact with external tools effectively.
\end{itemize}

A summary of representative agentic reasoning apporaches and their characteristics is provided in Table~\ref{tab:taxonomy}.
The following sections examine representative frameworks and techniques within each approach.

\subsection{Prompt-Based Approaches}


Prompt-based approaches harness the remarkable capabilities already present in pre-trained LLMs to enable agentic behavior. 
Instead of modifying the model's weights through training, these methods rely on sophisticated prompting techniques, few-shot examples or built-in tool interfaces, to guide the LLM in its interaction with external tools like search engines.

\textbf{Function-Calling-Based}: A foundational prompt-based method for agentic behavior, and one way to implement function calling, is ReAct (Reason+Act)  \cite{yao2023reactsynergizingreasoningacting}. ReAct aims to create a synergy between the reasoning processes and action-taking capabilities within an LLM. Its core mechanism involves prompting the LLM to generate outputs in an interleaved sequence of Thought, Action, and Observation. ReAct typically employs few-shot prompting, providing the LLM with examples that demonstrate this Thought-Action-Observation trajectory for solving similar tasks. These examples guide the frozen LLM on how to structure its reasoning, utilize available tools, and progress towards the goal. The framework demonstrated significant advantages, particularly in grounding the LLM's reasoning. By allowing the model to actively seek and incorporate external information via actions, ReAct can mitigate the hallucination and error propagation issues sometimes observed in purely internal reasoning methods like Chain-of-Thought (CoT) \cite{wei2023chainofthoughtpromptingelicitsreasoning}. 
The explicit reasoning traces (“Thoughts”) in ReAct enhance the interpretability and transparency of the model’s decision-making. Within RAG, ReAct offers a natural agentic reasoning pipeline: the LLM's "Thought" process can identify a knowledge gap, leading to a search "Action," with the retrieved results forming the "Observation" that informs subsequent reasoning.
A related method, Self-Ask~\cite{press2023measuringnarrowingcompositionalitygap}, encourages step-by-step problem decomposition by prompting the LLM to generate and answer simpler follow-up questions. These intermediate steps often involve search actions, enabling the model to gather relevant information before attempting to answer the main question.

Another prominent prompt-based approach involves leveraging the function calling or tool use capabilities that have been explicitly built into or fine-tuned into certain LLMs, such as versions of GPT \cite{openai_function_calling}, Llama, and Gemini. This feature allows the LLM to interact reliably with predefined external tools or APIs based on natural language instructions. Function calling significantly expands the capabilities of LLMs beyond text generation, enabling them to access real-time, dynamic information, interact with external systems and databases, automate tasks, and reliably convert natural language requests into structured API calls or database queries. In contrast to the more open-ended "thought-action-observation" cycle of ReAct, function calling often bypasses explicit intermediate reasoning steps. The LLM directly identifies the relevant tool and generates the necessary parameters based on its training to recognize and format specific function calls. This more direct approach relies on the model's pre-existing knowledge of available tools and their required inputs. Furthermore, the format and capabilities of the tools accessible via function calling are typically predefined and have been integrated into the model's training or prompt design. For Agentic RAG, function calling provides a straightforward and structured way for the LLM agent to invoke a search API when its internal analysis determines that external information is required to answer a prompt accurately.

\textbf{Large Reasoning Model-based}: A growing trend in Agentic RAG workflow involves directly utilizing LLMs that possess inherently strong reasoning capabilities, often referred to as Large Reasoning Models (LRMs). These models, sometimes developed through techniques like large-scale reinforcement learning (e.g., models analogous to OpenAI's o1 \cite{openai2024openaio1card}, DeepSeek-R1 \cite{deepseekai2025deepseekr1incentivizingreasoningcapability}), are designed to excel at complex, multi-step reasoning tasks. The underlying premise is that an LLM with superior intrinsic reasoning abilities will be better equipped to manage the complexities of an Agentic RAG workflow, including decomposing challenging queries, planning information-gathering steps, assessing the relevance and utility of retrieved information, and synthesizing knowledge effectively. In essence, leveraging LRMs within RAG represents a prompt-based agentic strategy where the model's powerful inherent reasoning capabilities drive the process, implicitly deciding when and how to retrieve information to support its complex thought processes. 
\begin{figure}[h]
	\centering
	\includegraphics[width=0.98\linewidth]{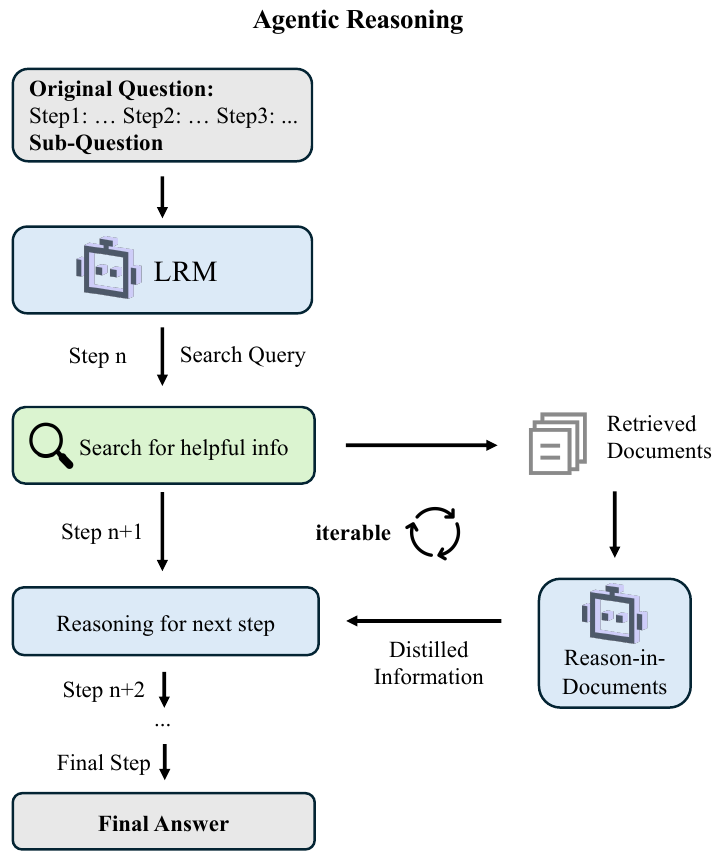}
	\caption{A demonstration of Agentic Reasoning.}
	\label{agentic reasoning figure} 
\end{figure}

However, effectively managing the retrieved context is another significant challenge. LLMs with extremely long context windows can suffer from a "lost-in-the-middle" problem, where information presented in the middle of a long input receives less attention. 
Furthermore, retrieved documents, whether in long-context models or standard RAG, often contain verbose, noisy or contradictory content that can disrupt the coherence of the LLM's reasoning process.
Mitigating this challenge requires more precise retrieval strategies and adaptive context management mechanisms.
The Search-o1 framework \cite{li2025searcho1agenticsearchenhancedlarge} is specifically designed to enhance LRMs by tackling knowledge insufficiency during long, step-by-step reasoning chains. It integrates two core components: an Agentic RAG Mechanism where the LRM dynamically triggers search queries based on self-assessed knowledge gaps, and a Reason-in-Documents Module that processes retrieved content to dist ill relevant information into a refined format, thereby minimizing noise and maintaining the LRM's reasoning integrity. Search-o1 exemplifies a sophisticated prompt-based agentic approach focused on maintaining reasoning integrity in the face of external information retrieval.

\subsection{Training-Based Approaches}


While prompt-based methods leverage the inherent capabilities of LLMs, their performance in complex tool-use scenarios can be inconsistent. Achieving highly reliable and optimized behavior, especially in deciding when and how to interact with tools like search engines, often benefits from explicit training. Training-based approaches, particularly those utilizing Reinforcement Learning (RL), enable the LLM agent to learn sophisticated strategies through trial and error, directly optimizing its actions towards specific goals such as maximizing retrieval effectiveness or overall task success. RL enables agents to develop more robust and strategic interaction patterns than prompting alone. 

\textbf{Interacting with local retrieval systems}: Search-R1 \cite{jin2025searchr1trainingllmsreason} tackles a different aspect of agentic search: training the LLM to autonomously decide when to search and what to search for during a multi-step reasoning process. It extends RL-based reasoning frameworks (like DeepSeek-R1) by integrating search engine interaction directly into the learning loop. In the Search-R1 framework, the search engine is modeled as part of the RL environment. The LLM agent learns a policy to generate a sequence of tokens that includes both internal reasoning steps (often enclosed in \texttt{<think>} tags) and explicit triggers for search actions. These triggers are special tokens, \texttt{<search>} and \texttt{</search>}, which encapsulate the generated search query. This design allows for flexible, multi-turn interactions where the LLM can interleave reasoning, searching, processing retrieved information (presented within \texttt{<information>} tags), and further reasoning or searching as needed. The framework utilizes a simple outcome-based reward function, typically based on the correctness of the final answer generated by the LLM (within \texttt{<answer>} tags) compared to a ground truth, avoiding the complexity of designing intermediate process rewards. A crucial technique employed is retrieved token masking. During the calculation of the RL loss (using algorithms like PPO or GRPO \cite{shao2024deepseekmathpushinglimitsmathematical}), the tokens corresponding to the content retrieved from the search engine (i.e., within the \texttt{<information>} tags) are ignored or masked out, which stabilizes the training process. Search-R1 has shown significant performance improvements over various RAG baselines on question-answering datasets. Its core contribution is training the LLM to learn an optimal policy for interacting with the search engine as an integrated part of its reasoning flow, enabling dynamic, context-aware search decisions. The related R1-Searcher \cite{song2025r1searcherincentivizingsearchcapability} framework also proposes a similar two-stage, outcome-based RL approach for enhancing search capabilities.



ReZero (Retry-Zero) \cite{dao2025rezeroenhancingllmsearch} introduces another dimension to RL-based agentic search by specifically focusing on incentivizing persistence. It addresses the common scenario where an initial search query might fail to retrieve the necessary information, potentially causing the LLM agent to halt prematurely or generate a suboptimal response. ReZero aims to teach the agent the value of ``trying one more time." The framework operates within a standard RL setup (using GRPO is mentioned) where the LLM interacts with a search environment. The novelty lies in its modified reward function, which includes a specific component termed reward retry. This component provides a positive reward signal whenever the LLM issues a \texttt{<search>} query after the initial search query within the same reasoning trajectory. Crucially, this reward for retrying is conditional upon the agent successfully completing the task, indicated by generating a final answer enclosed in \texttt{<answer>} tags. This conditionality prevents the agent from accumulating rewards simply by retrying indefinitely without making progress. By directly rewarding the act of persistence (when productive), ReZero encourages the LLM to explore alternative queries or search strategies if the first attempt proves insufficient. This contrasts with methods that might only implicitly reward persistence through eventual task success. ReZero positions itself as complementary to frameworks like DeepRetrieval; 
while DeepRetrieval focuses on optimizing a single refined query, ReZero emphasizes the value of making multiple retrieval attempts when needed.

\textbf{Interacting with real-world search engines}:
DeepRetrieval \cite{jiang2025deepretrievalhackingrealsearch} focuses specifically on improving the quality of the search queries generated by the LLM agent. It frames the task of query generation or rewriting as an RL problem, training the LLM to transform an initial user query into a more effective query for downstream retrieval systems. The core mechanism involves the LLM generating an augmented or rewritten query based on the input query. DeepRetrieval employs RL algorithms like Proximal Policy Optimization (PPO) \cite{schulman2017proximalpolicyoptimizationalgorithms} to train this query generation process. A key innovation lies in its reward signal: instead of relying on supervised data (e.g., pairs of original and "gold" rewritten queries), DeepRetrieval uses the performance of the generated query in the actual retrieval system as the reward. Metrics such as recall@k, Normalized Discounted Cumulative Gain (NDCG), or evidence-seeking retrieval accuracy (Hits@N) obtained from executing the generated query against a restricted real search engine (like PubMed) or document collection are used to provide feedback to the LLM. The model learns, through trial and error, to generate queries that maximize these retrieval metrics. To structure the generation, the model often produces reasoning steps within \texttt{<think>} tags before outputting the final query in an \texttt{<answer>} tag. This approach offers significant advantages. By directly optimizing for the end goal (retrieval performance), it bypasses the need for expensive and potentially suboptimal supervised query datasets. Compared to other RL methods, DeepRetrieval's primary focus is on optimizing the content and formulation of the search query itself.

DeepResearcher \cite{zheng2025deepresearcherscalingdeepresearch} pushes the boundaries of training-based Agentic RAG by moving beyond controlled environments or static corpora to perform end-to-end RL training directly within real-world web environments. It aims to equip LLM agents with the capabilities needed for complex, deep research tasks that require navigating the noisy, unstructured, and dynamic nature of the open web. This addresses a key limitation of many existing agents, whether prompt-engineered or trained in simulated/static RAG settings, which often struggle with the complexities of real-world web interaction. The framework employs RL (specifically GRPO with an F1 score-based reward for answer accuracy ) to train agents that interact with live web search APIs and browse actual webpages. DeepResearcher utilizes a specialized multi-agent architecture to handle the complexities of web interaction. This includes a reasoning module, a tool for invoking web search, and dedicated ``browsing agents" responsible for extracting relevant information from the diverse structures of webpages encountered. Training in this realistic setting was found to foster several emergent cognitive behaviors not typically observed in agents trained under more constrained conditions. These include the ability to formulate initial plans and dynamically adjust them during the research process, cross-validate information retrieved from multiple web sources, engage in self-reflection when retrieved information seems contradictory or insufficient leading to refined search strategies, and exhibit honesty by declining to provide an answer when definitive information cannot be found. DeepResearcher demonstrated substantial performance improvements over prompt-engineering baselines and RAG-based RL agents trained on static corpora, particularly on open-domain research tasks. The results strongly suggest that end-to-end training in realistic web environments is crucial for developing robust and capable research agents, moving closer to the capabilities hinted at by proprietary systems like OpenAI's Deep Research \cite{OpenAI2025deepresearch} or Grok's DeeperSearch.

The progression for the training-based methods, from optimizing the decision process of when and what to query (Search-R1), to fostering persistence (ReZero), optimizing query formulation (DeepRetrieval), and managing real-world research workflows (DeepResearcher) reflects the growing sophistication of RL in agentic search.
It reflects a growing appreciation that effective information seeking by an agent involves a confluence of factors: query quality, strategic timing, resilience to failure, and adeptness in navigating realistic information environments and so on. Future advancements in RL-based Agentic RAG will likely need to integrate these facets more holistically, perhaps through more complex reward structures, multi-objective optimization, or architectures that explicitly model these different dimensions of the search process, to achieve truly human-like research and problem-solving capabilities.

\section{Future Research Directions}
\label{future}
\textbf{Enhancing tool interaction through advanced configuration.} 
Current agentic reasoning often utilizes search tools with relatively basic interfaces, primarily focused on generating text queries. Future work should enable agents to exploit more advanced configurations offered by external APIs and tools. This could involve training agents to understand and utilize options like result filtering (e.g., by date, source type), sorting criteria, specifying search domains, or interacting with structured databases via complex queries. Granting finer control would support more targeted, efficient, and strategic retrieval aligned with task demands.


\textbf{Developing finer-Grained and process-oriented reward functions.} 
Simple outcome-based rewards like exact match may not offer adequate guidance for complex RAG tasks that require multi-step reasoning or detailed responses. Future research should develop fine-grained reward functions that assess both final answer correctness and intermediate steps such as document relevance, reasoning coherence, information cross-validation, and effective problem decomposition. These signals are vital for training agents to handle queries that demand more than short factual answers.

\textbf{Improving Efficiency in Retrieval.}
The approaches mentioned above primarily focus on the accuracy of the final answer, but enhancing the efficiency of the retrieval process itself is also critical. Agents trained to interact with potentially vast information sources, must learn to perform retrievals strategically. Future research should focus on techniques that help agents avoid excessive or unnecessary search queries, select the most promising sources, and know when sufficient information has been gathered. Developing strategies to prevent agents from getting stuck in loops of unproductive searching or performing redundant retrievals is vital for practical and scalable Agentic RAG.

\textbf{Enhancing Generalization and Robustness in Dynamic Environments.}
Robust generalization to new queries, unseen tools (e.g., sparse, dense, or web retrieval), and changing environments remains a major challenge. While training in realistic conditions (as in DeepResearcher) improves resilience, agents still struggle with tool failures or shifting knowledge availability. Future work should explore adaptive training methodologies and architectures that ensure robust performance in unfamiliar or dynamic settings.



By addressing key areas such as improving agent control over tools, designing more sophisticated reward signals, increasing efficiency, and enhancing generalization, the field can move toward building more capable, reliable, and widely applicable Agentic RAG systems. These advancements are essential for transitioning agentic AI from research prototypes to practical systems that can effectively support humans in complex information tasks.


\section{Conclusions}
\label{conculsion}
As language models are increasingly deployed in complex, knowledge-intensive applications, the limitations of static RAG pipelines have become apparent. Reasoning Agentic RAG offers a promising path forward by integrating retrieval with model-driven planning, self-reflection, and tool use. This paper surveyed the landscape of reasoning workflows within Agentic RAG, distinguishing between predefined reasoning with fixed orchestration, and agentic reasoning that enables dynamic, autonomous decision-making. We reviewed key methods across both paradigms, highlighting their strengths, limitations, and use-case applicability. 
To advance the field, we identify several crucial directions for future research, including fine-grained reward design, enhanced tool control, automated data synthesis, and robust training in dynamic environments.
These innovations will be essential for realizing intelligent, context-aware RAG systems capable of addressing real-world challenges with greater adaptability, transparency, and reliability. 



\bibliographystyle{named}
\bibliography{main}

\end{document}